\documentclass{article}

\usepackage{arxiv}

\usepackage[utf8]{inputenc} 
\usepackage[T1]{fontenc}    
\usepackage{hyperref}       
\usepackage{url}            
\usepackage{booktabs}       
\usepackage{amsfonts}       
\usepackage{nicefrac}       
\usepackage{microtype}      
\usepackage{lipsum}		
\usepackage{graphicx}
\usepackage{doi}

\usepackage{booktabs}
\usepackage{multirow}
\usepackage{graphicx}
\usepackage{cite}
\usepackage{lipsum}
\usepackage[table,xcdraw]{xcolor}
\usepackage{amssymb}
\usepackage{amsmath,amssymb,amsfonts}
\usepackage{lscape}
\usepackage{appendix}

\title{Robotic and Generative Adversarial Attacks in Offline Writer-independent Signature Verification}

\author{ \href{https://orcid.org/0000-0002-9858-1231}{\includegraphics[scale=0.06]{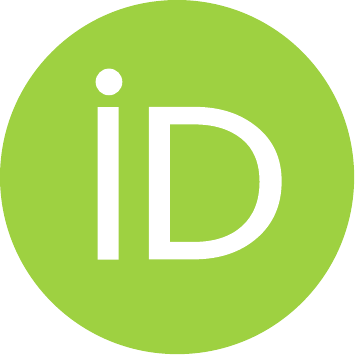}\hspace{1mm}Jordan J.~Bird}\thanks{\url{https://jordanjamesbird.com/}} \\
	Computational Intelligence and Applications Research Group (CIA) \\
	Department of Computer Science\\
	Nottingham Trent University\\
	Nottingham, United Kingdom \\
	\texttt{jordan.bird@ntu.ac.uk} \\
}

\renewcommand{\shorttitle}{Robotic and Generative Adversarial Attacks in Signature Verification}

\hypersetup{
pdftitle={Improving Customer Service Chatbots with Attention-based Transfer Learning},
pdfauthor={Jordan J.~Bird},
pdfkeywords={Chatbot, Robotics, Human-Robot Interaction, Social Robotics, Natural Language Processing, Transformers, Deep Learning},
}

\begin{document}
\maketitle

\begin{abstract}
This study explores how robots and generative approaches can be used to mount successful false-acceptance adversarial attacks on signature verification systems. Initially, a convolutional neural network topology and data augmentation strategy are explored and tuned, producing an 87.12\% accurate model for the verification of 2,640 human signatures. Two robots are then tasked with forging 50 signatures, where 25 are used for the verification attack, and the remaining 25 are used for tuning of the model to defend against them. Adversarial attacks on the system show that there exists an information security risk; the Line-us robotic arm can fool the system 24\% of the time and the iDraw 2.0 robot 32\% of the time. A conditional GAN finds similar success, with around 30\% forged signatures misclassified as genuine. Following fine-tune transfer learning of robotic and generative data, adversarial attacks are reduced below the model threshold by both robots and the GAN. It is observed that tuning the model reduces the risk of attack by robots to 8\% and 12\%, and that conditional generative adversarial attacks can be reduced to 4\% when 25 images are presented and 5\% when 1000 images are presented. 
\end{abstract}


\section{Introduction}
To forge a signature with the aim of deceiving is a serious crime throughout the world\cite{cowley1983forgery,hemraj2002crime} with financial implications and risks to personal identity. There are countless examples of signature forgery and the issues it causes such as fake historical documents fraudulently signed with Abraham Lincoln's signature\cite{nickell2005detecting}, the forgery and subsequent sale of celebrity signatures\cite{mcmillen2002sports}, as well as the forging and cashing of cheques\cite{farnsworth1960insurance}. To have a signature verified by a human expert is an expensive endeavour, and is thus oftentimes another step in the process which bares even further financial implications for an individual or a business. In the UK alone, cheque fraud resulted in losses of £12.3 million in 2020 following £53.6 million losses in 2019\cite{uk_finance_2021}. The COVID-19 lockdown introduced a level of fraud that had not been seen for several years. In this study, losses of £558.8 million were also shown to be prevented from cheque fraud in 2019. According to the American Bankers Association, cheque fraud amounted to \$15.1 billion in 2018 and affected around half a million individuals\cite{trentmann_2020}.

Modern computing proposes several solutions to the detection of forged signatures through autonomous verification as an added layer of protection against such attempts. Much state-of-the-art work in signature verification is to detect when a human being has forged another's signature. Given the rapid growth of consumer robotics due to their ease of use and low cost, the level of detail now possible via robotic signature forgery is a growing concern. Rapid analysis and near-perfect replication of a signature is now possible by machines that cost a fraction of the price of the average smartphone. In terms of generative approaches, it is possible to train models such as GANs on home computers; that is, with hardware already found in the home, signature verification systems can be succesfully attacked and fooled by data generated by a neural network. In this study, we explore how robots and generative adversarial methods can be used to fool offline writer-independent signature verification systems. Following this, efforts are made to tune and improve such systems to provide a preliminary line of defence against such adversarial attacks. 

The multiple scientific contributions presented by this work are as follows: (i) tuning of a vision-based system inspired by the current State-of-the-Art for accurate signature verification. (ii) a pipeline to analyse signatures (raster images), produce vectors, and then G-code for execution of two pen-holding robotic arms (Line-us and iDraw 2.0). (iii) a Conditional Generative Adversarial Network to discern and generate real and forged signatures. (iv) Successful adversarial attacks on the verification system by both robots and generative approaches. (v) Successful defence of the verification system by fine-tune transfer learning from examples produced during adversarial attack. To the author's knowledge, this article proposes the first case of an attack on a signature verification system by using robots to physically copy and write signatures. 

The remainder of this article is as follows: Section \ref{sec:background} reviews the background of the field, including the state of the art in signature verification and adversarial attacks. Following this, Section \ref{sec:method} outlines the method followed by the experiments in this work, and the results are presented and discussed in Section \ref{sec:method}. Finally, concluding remarks and future work are discussed in Section \ref{sec:conclusion}.

\section{Background}
\label{sec:background}
This section explores the current state of the art related to this work. This includes biometrics, signature verification with visual and deep learning approaches, and methods of attacking verification methods. 

Biometrics are systems that recognise an individual based on a given input. For example, recognition of an individual based on their fingerprint\cite{praseetha2020secure}, speech patterns\cite{boles2017voice,bird2020overcoming}, EEG\cite{maiorana2021learning} and ECG\cite{ibtehaz2021edith} signals, or the iris of their eye\cite{al2018multi} to name a few. Signature verification is a biometric for the recognition and verification of an individual based on the way that they sign their name\cite{houmani2012biosecure}. Given the ability that a genuine signature has, successful forgeries can therefore have major implications related to personal identity and finances. Online signature verification deals with smart devices, such as tablets and pens, that record the resultant signature along with features such as pressure, azimuth, velocity, and inclination\cite{soelistio2021review}. Offline signature verification is based on the resultant signature alone, and is more common given that signed paper documents are often signed. 

Earlier works such as Kalera et al.\cite{kalera2004offline} proposed methods such as distance metrics between examples of real and forged signatures. The identification of two data sets was found to reach an accuracy of around 93\%, while verification was possible for around 78\% of the data objects in one dataset and around 68\% in the second set. In \cite{ferrer2012robustness}, the authors proposed a machine learning approach for least-squares support vector machines with respect to the grey-level features observed. Consideration of grey levels within pixel values provides pseudo-online insight into pen velocity and pressure. Such features are noted to be important given that a human will sign their own signature quickly, while a forgery is oftentimes more thought-out and therefore takes longer to replicate. 

State-of-the-art research on signature verification relies largely on deep learning-based approaches. In 2019, Sam et al.\cite{sam2019offline} proposed an offline verification technique with the Inception-v3 Convolutional Neural Network (CNN) that achieved 88\% validation accuracy on 170 images after training on 370 signatures. A similar approach was explored in \cite{yapici2018convolutional}, with a CNN model achieving 62.5\% writer-independent and 75\% writer-dependent signature verification. In this study, the writer-independent dataset contained 300 images for training and 240 for validation, and the writer-dependent dataset contained 30 images for training and 24 for validation. In 2018, Souza et al.\cite{souza2018writer} proposed a feature extraction technique with a CNN prior to classification by Support Vector Machines; achieving an equal error rate of 1.48\% on the Brazilian PUC-PR dataset.

Given that there is a growing reliance on automated systems for verification, therein lies the risk of attack. An adversarial attack is a method of fooling a machine learning model by engineering input data to force a chosen output prediction. In signature verification, a successful attack would allow for an individual to pass another's signature without their participation or knowledge. Work by the Autonomous University of Madrid's Biometric Recognition Group showed that a Bayesian hill-climbing attack could overcome a signature verification system 95\% of the time\cite{galbally2007bayesian}. Similarly, Li et al.\cite{li2021black} proposed generating invisible perturbations on signatures which led to a verification system being succesfully attacked in 92.1\% of cases. In \cite{galbally2009evaluation} the authors proposed to generate synthetic signatures by spectral analysis of trajectory functions, presenting results of up to 0.04\% brute-force attack successes. Scheidat et al. \cite{scheidat2005distance} proposed a strategy of distance-level fusion to overcome brute-force attacks on verification systems, noting a lower error rate when combining online verification experts into a unified system. In \cite{buccafurri2009fortifying}, an attack is described which can implement digital signatures into documents, by embedding PDF and TIFF files within a file which goes largely undetected. In another related study, Alonso-Fernandez et al.\cite{alonso2009robustness} described how forgeries do not only simply attack a system while remaining static, rather, forger skills are improved over time. This is similar to a generative adversarial learning framework, which inspired the added approach to this study alongside the two robots.

\section{Method}
\label{sec:method}
\begin{figure}
    \centering
    \includegraphics[scale=0.9]{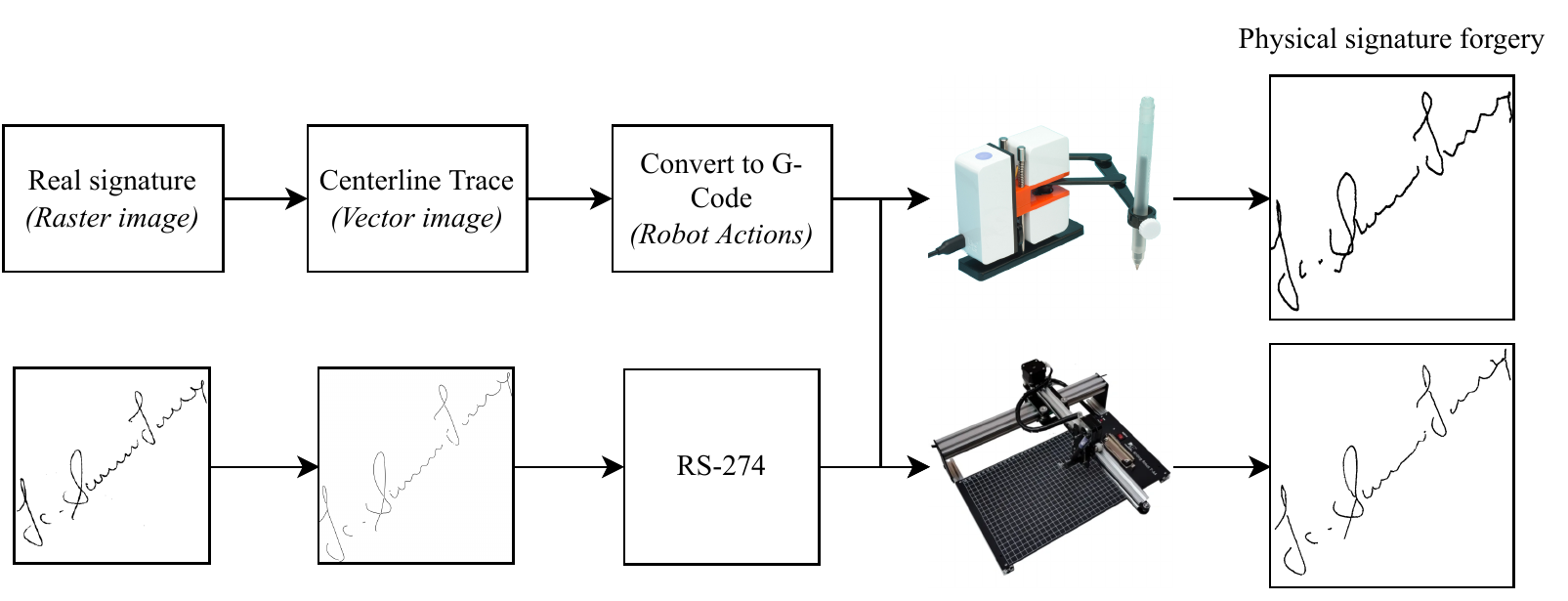}
    \caption{The general approach for robotic signature forgery. The raster image is converted to a centerline vector, which is then in turn converted to G-Code. The robots then execute the code and physically forge the given signature with a pen. The outputs of this diagram are the paper signatures scanned and pre-processed.}
    \label{fig:general-diagram}
\end{figure}
In this section, the method followed by the proposed approach is described. Initially, data pre-processing and augmentation are considered, followed by topology engineering and learning for signature verification models. Following this, adversarial attacks and defences are then outlined. 

The data used in this study is the CEDAR Signature Verification Dataset\footnote{https://cedar.buffalo.edu/signature/} from \cite{srihari2008machine}. In the study, 55 individuals contributed 24 of their own signatures, and some of the participants forged the signatures of others. The dataset comprises 1,320 genuine and 1,320 forged signatures. 
\begin{figure}
    \centering
    \includegraphics[scale=0.9]{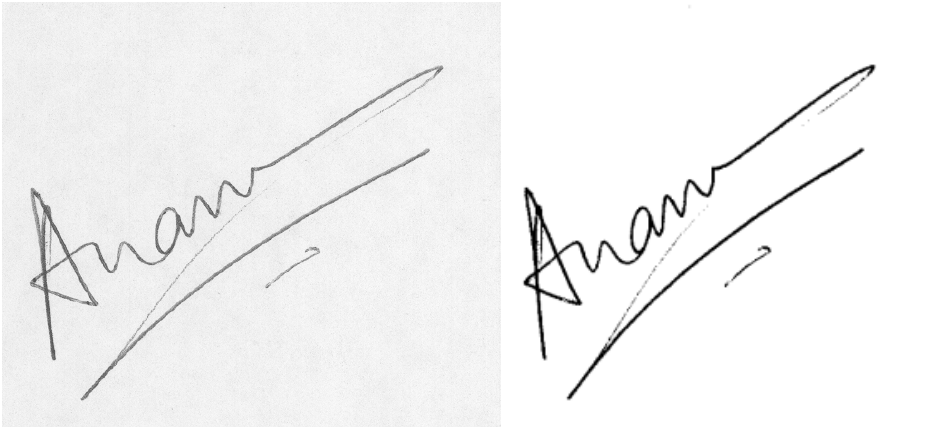}
    \caption{An example of a signature before (left) and after (right) preprocessing.}
    \label{fig:preprocessing-example}
\end{figure}
The dataset was found to contain images that varied in size and aspect ratio. Given the nature of the input topologies of convolutional neural networks, data pre-processing was performed. All images were resized to 256px and binarized; An example of an image before and after processing can be found in Figure \ref{fig:preprocessing-example}. Some signatures were also observed to be signed at a rotation of up to approximately 20\%. Random rotation is therefore explored as a potential method of data augmentation along with a vanilla convolutional neural network. 
Convolutional Neural Networks were the main learning method within the experiments. During an 80/20 split, topology tuning was performed via CNN layers of \{1, 2, 3\} in number, each containing \{16, 32, 64\} filters. Dense interpretation multilayer perceptrons were then tuned, involving 1 or 2 layers with \{64, 128, 256\} rectified linear units. Since some methods would likely converge sooner than others, the CNNs are therefore given an infinite number of epochs to train, instead stopping when there were no observed improvements on the validation metrics within a 25 epoch window. 
The Conditional GAN approach to generating signature forgeries was conditioned on the class label. The GAN generator began with an $8\times8$ random noise matrix before 5 transpose layers of 128 filters each upscaled the matrix to a $256\times256$ image. Each generator layer had a Leaky Rectified Linear Unit activation function. The discriminator network consisted of two convolutional layers with 128 filters and a 2,2 stride. A dropout value of 0.4 was implemented prior to output to prevent mode collapse. Adversarial attacks are then tested with 25 and 1000 images output by the GAN, conditioned with a class label corresponding to the real signature class.
\begin{figure}
    \centering
    \includegraphics{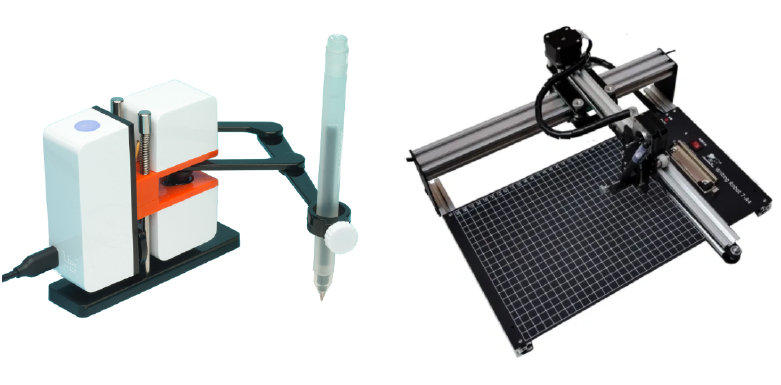}
    \caption{The two robots used in this study; Line-us (left) and iDraw 2.0 (right).}
    \label{fig:robots-images}
\end{figure}

In addition to the generative adversarial attack, this study also explores whether robotic arms can forge signatures. Two robots are used, the Line-us robotic arm and the iDraw 2.0 handwriting plotter. These two robots can be seen in Figure \ref{fig:robots-images}. Each robot was tasked with replicating the same 25 signatures from the dataset, and these images provide robotic attacks on the verification model. Figure \ref{fig:general-diagram} shows the general approach to enabling physical replication of the signature; The raster scan is vectorised through a centerline trace, which is then converted to robot-compatible G-Code. The G-Code is executed on the robots, which then write the signature on paper.

For defence, a further 25 real signatures are signed by the two robots (and generated by the GAN). The model is fine-tuned for one epoch, with the resultant signatures included within the dataset labelled as forgeries. Therefore, this strategy exposes the verification model to robot-forged signatures with the aim of providing further enhanced awareness. The aim thus is to classify behaviours that the GAN and robots exhibit which humans necessarily do not. 

\section{Results}
\label{sec:results}
In this section, the results of the proposed approach are presented. The verification model is first tuned and forgery methods are discussed. Following this, the robotic and generative adversarial attacks are performed and defended against with fine-tune transfer learning.
\begin{table}[]
\centering
\caption{Topology tuning for signature verification without data augmentation.}
\label{tab:results-aug}
\begin{tabular}{@{}rrrrrrr@{}}
\toprule
\multicolumn{1}{l}{\textbf{\begin{tabular}[c]{@{}l@{}}CNN \\ Layers\end{tabular}}} & \multicolumn{1}{l}{\textbf{\# Filters}} & \multicolumn{1}{l}{\textbf{Max Epoch}} & \multicolumn{1}{l}{\textbf{Acc.}} & \multicolumn{1}{l}{\textbf{Prec.}} & \multicolumn{1}{l}{\textbf{Rec.}} & \multicolumn{1}{l}{\textbf{F1}} \\ \midrule
\textit{\textbf{1}}                                                                & \textit{\textbf{16}}                    & 41                                     & 75.38                             & 0.75                               & 0.761                             & 0.756                           \\
\textit{\textbf{2}}                                                                & \textit{\textbf{16}}                    & 52                                     & 78.41                             & 0.791                              & 0.773                             & 0.782                           \\
\textit{\textbf{3}}                                                                & \textit{\textbf{16}}                    & 52                                     & 81.63                             & 0.82                               & 0.811                             & 0.815                           \\
\textit{\textbf{1}}                                                                & \textit{\textbf{32}}                    & 31                                     & 73.3                              & 0.734                              & 0.731                             & 0.732                           \\
\textit{\textbf{2}}                                                                & \textit{\textbf{32}}                    & 35                                     & 77.84                             & 0.78                               & 0.777                             & 0.778                           \\
\textit{\textbf{3}}                                                                & \textit{\textbf{32}}                    & 47                                     & 82.39                             & 0.833                              & 0.811                             & 0.822                           \\
\textit{\textbf{1}}                                                                & \textit{\textbf{64}}                    & 28                                     & 74.81                             & 0.73                               & 0.788                             & 0.758                           \\
\textit{\textbf{2}}                                                                & \textit{\textbf{64}}                    & 29                                     & 78.79                             & 0.82                               & 0.739                             & 0.777                           \\
\textit{\textbf{3}}                                                                & \textit{\textbf{64}}                    & 43                                     & 84.09                             & 0.838                              & 0.845                             & 0.842                           \\
\multicolumn{1}{l}{\textbf{Mean}}                                                  & \multicolumn{1}{l}{}                    & 39.78                                  & 78.52                             & 0.79                               & 0.78                              & 0.78                            \\ \bottomrule
\end{tabular}
\end{table}
\begin{table}[]
\centering
\caption{Topology tuning for signature verification with data augmentation (random rotation, max. 20\%).}
\label{tab:results-no-aug}
\begin{tabular}{@{}rrrrrrr@{}}
\toprule
\multicolumn{1}{l}{\textbf{\begin{tabular}[c]{@{}l@{}}CNN \\ Layers\end{tabular}}} & \multicolumn{1}{l}{\textbf{\# Filters}} & \multicolumn{1}{l}{\textbf{Max Epoch}} & \multicolumn{1}{l}{\textbf{Acc.}} & \multicolumn{1}{l}{\textbf{Prec.}} & \multicolumn{1}{l}{\textbf{Rec.}} & \multicolumn{1}{l}{\textbf{F1}} \\ \midrule
\textit{\textbf{1}}                                                                & \textit{\textbf{16}}                    & 96                                     & 80.11                             & 0.777                              & 0.845                             & 0.809                           \\
\textit{\textbf{2}}                                                                & \textit{\textbf{16}}                    & 75                                     & 85.04                             & 0.832                              & 0.879                             & 0.855                           \\
\textit{\textbf{3}}                                                                & \textit{\textbf{16}}                    & 83                                     & 85.04                             & 0.829                              & 0.883                             & 0.855                           \\
\textit{\textbf{1}}                                                                & \textit{\textbf{32}}                    & 99                                     & 78.03                             & 0.766                              & 0.807                             & 0.786                           \\
\textit{\textbf{2}}                                                                & \textit{\textbf{32}}                    & 88                                     & 87.12                             & 0.855                              & 0.894                             & 0.874                           \\
\textit{\textbf{3}}                                                                & \textit{\textbf{32}}                    & 99                                     & 87.12                             & 0.926                              & 0.807                             & 0.862                           \\
\textit{\textbf{1}}                                                                & \textit{\textbf{64}}                    & 115                                    & 81.06                             & 0.793                              & 0.841                             & 0.816                           \\
\textit{\textbf{2}}                                                                & \textit{\textbf{64}}                    & 98                                     & 86.55                             & 0.846                              & 0.894                             & 0.87                            \\
\textit{\textbf{3}}                                                                & \textit{\textbf{64}}                    & 60                                     & 83.9                              & 0.833                              & 0.849                             & 0.841                           \\
\multicolumn{1}{l}{\textbf{Mean}}                                                  & \multicolumn{1}{l}{}                    & 90.33                                  & 83.77                             & 0.83                               & 0.86                              & 0.84                            \\ \bottomrule
\end{tabular}
\end{table}
\begin{table}[]
\centering
\caption{Interpretation network topology tuning with data augmentation.}
\label{tab:results-dense}
\begin{tabular}{rrrrrrr}
\hline
\multicolumn{1}{l}{\textbf{\begin{tabular}[c]{@{}l@{}}MLP \\ Layers\end{tabular}}} & \multicolumn{1}{l}{\textbf{\begin{tabular}[c]{@{}l@{}}MLP \\ Neurons\end{tabular}}} & \multicolumn{1}{l}{\textbf{\begin{tabular}[c]{@{}l@{}}Max \\ Epoch\end{tabular}}} & \multicolumn{1}{l}{\textbf{Acc.}} & \multicolumn{1}{l}{\textbf{Prec.}} & \multicolumn{1}{l}{\textbf{Rec.}} & \multicolumn{1}{l}{\textbf{F1}} \\ \hline
\textit{\textbf{1}}                                                                & \textit{\textbf{16}}                                                                & 53                                                                                & 80.3                              & 0.808                              & 0.796                             & 0.802                           \\
\textit{\textbf{2}}                                                                & \textit{\textbf{16}}                                                                & 26                                                                                & 50                                & \multicolumn{1}{l}{-}              & \multicolumn{1}{l}{-}             & \multicolumn{1}{l}{-}           \\
\textit{\textbf{1}}                                                                & \textit{\textbf{32}}                                                                & 74                                                                                & 82.58                             & 0.812                              & 0.849                             & 0.83                            \\
\textit{\textbf{2}}                                                                & \textit{\textbf{32}}                                                                & 64                                                                                & 81.63                             & 0.808                              & 0.83                              & 0.819                           \\
\textit{\textbf{1}}                                                                & \textit{\textbf{64}}                                                                & 80                                                                                & 84.28                             & 0.842                              & 0.845                             & 0.843                           \\
\textit{\textbf{2}}                                                                & \textit{\textbf{64}}                                                                & 122                                                                               & 85.23                             & 0.832                              & 0.883                             & 0.857                           \\
\textit{\textbf{1}}                                                                & \textit{\textbf{128}}                                                               & 88                                                                                & 87.12                             & 0.855                              & 0.894                             & 0.874                           \\
\textit{\textbf{2}}                                                                & \textit{\textbf{128}}                                                               & 86                                                                                & 86.55                             & 0.861                              & 0.871                             & 0.866                           \\
\textit{\textbf{1}}                                                                & \textit{\textbf{256}}                                                               & 101                                                                               & 85.8                              & 0.829                              & 0.9                               & 0.864                           \\
\textit{\textbf{2}}                                                                & \textit{\textbf{256}}                                                               & 80                                                                                & 87.5                              & 0.856                              & 0.9                               & 0.878                           \\ \hline
\end{tabular}
\end{table}
Initially, Tables \ref{tab:results-aug} and \ref{tab:results-no-aug} present results for topology tuning with and without data augmentation, respectively. On average, augmentation led to a mean increase in accuracy by 5.25\%, but required a mean of 50.55 additional epochs to meet the early stopping criteria. The best convolutional models for signature verification in terms of classification accuracy (87.12\%) were found to be those with 2 and 3 convolutional layers, both with 32 filters. In terms of metrics, 32 filters within 2 layers were slightly more stable. As can be observed from Table \ref{tab:results-dense}, other dense interpretation networks were also tested; leading to an overall best score of 87.12\%. 

\begin{figure}
    \centering
    \includegraphics[scale=0.65]{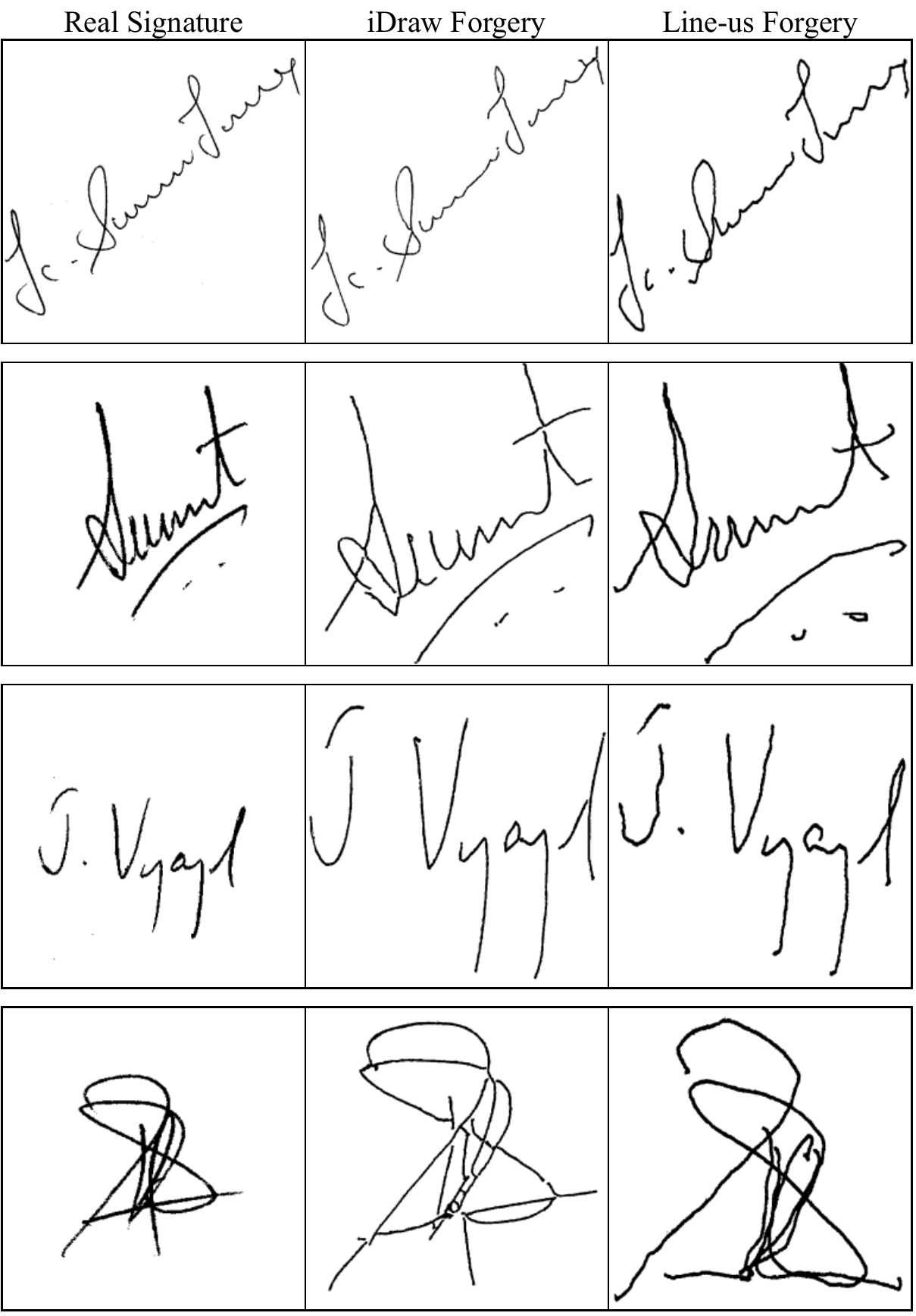}
    \caption{Real signatures compared to the forgeries by the iDraw and Line-us robots. Forgeries have been written physically before being scanned and pre-processed.}
    \label{fig:comparison}
\end{figure}

Figure \ref{fig:comparison} shows a real signature and the two counterparts forged by the iDraw 2.0 and Line-us robots. It can be observed that the iDraw 2.0 arguably creates the most realistic forgery, but features such as line intersections can differentiate it from the genuine image. Both robots seem to lose some detail, seen here via the minute squiggles in the line within the middle name; though this is not necessarily indicative of forgery given that human beings will also change the level of detail given due to reasons such as situationally dependent writing speed. 

\begin{figure}
    \centering
    \includegraphics[scale=0.6]{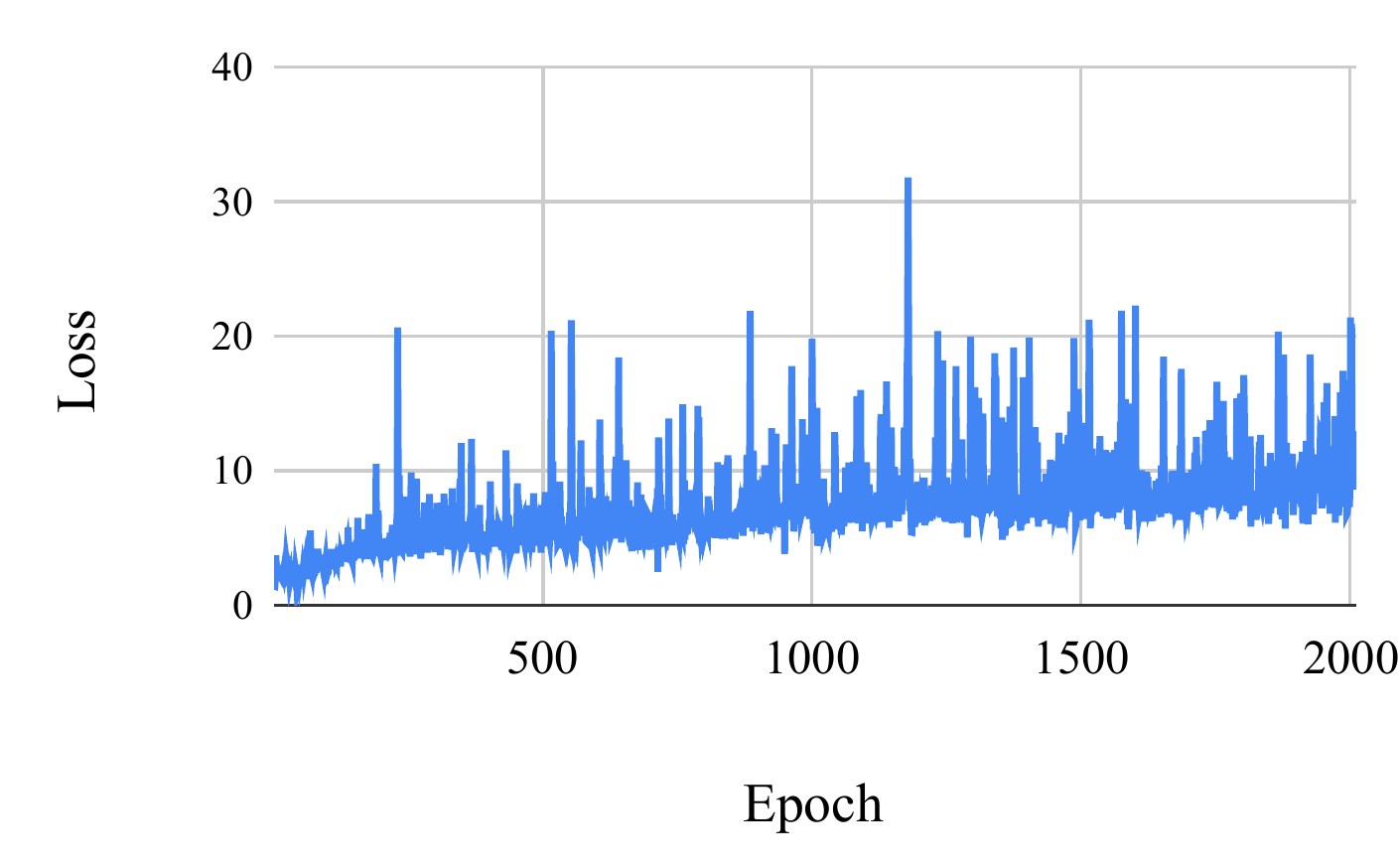}
    \caption{Generator loss per epoch for the Conditional Generative Adversarial Network.}
    \label{fig:gen_chart}
\end{figure}
\begin{figure}
    \centering
    \includegraphics[scale=0.6]{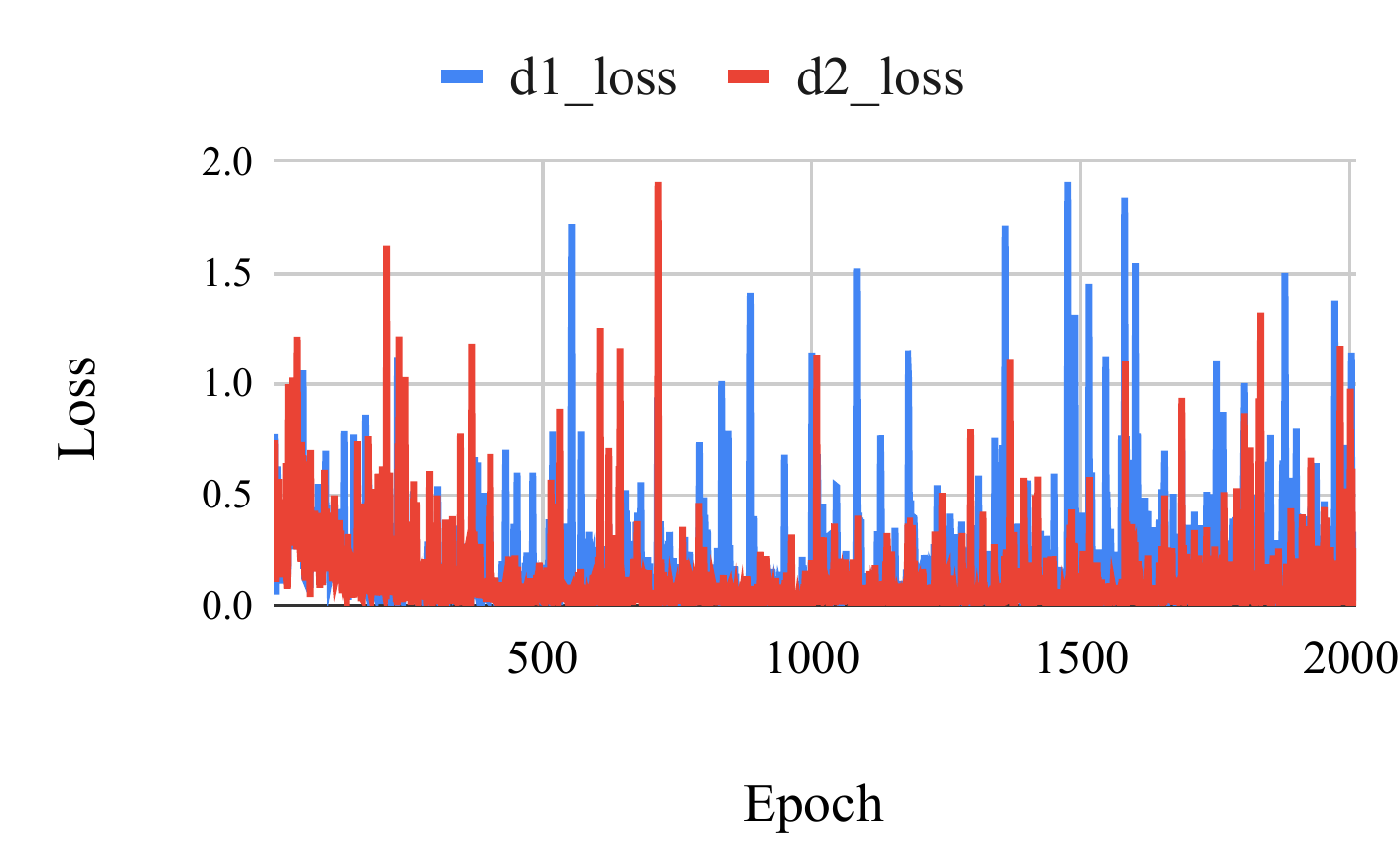}
    \caption{Discriminator losses per epoch for the Conditional Generative Adversarial Network.}
    \label{fig:disc_chart}
\end{figure}
\begin{figure}
    \centering
    \includegraphics[scale=0.65]{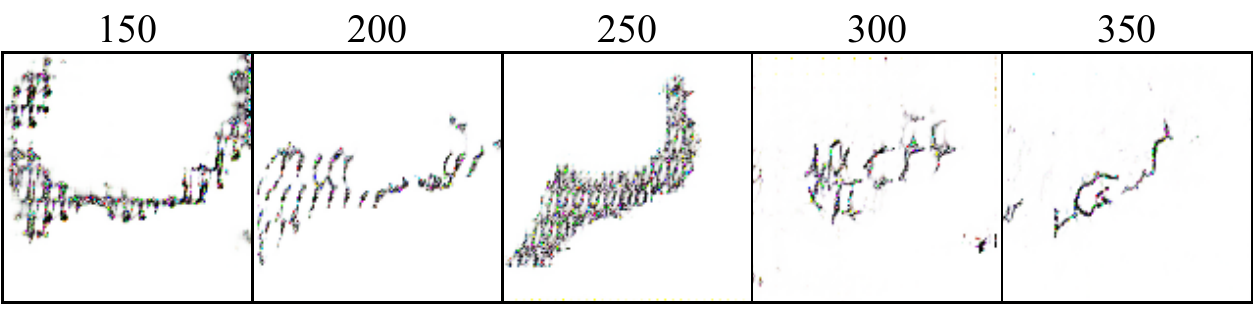}
    \caption{Early per-epoch examples of Conditional GAN samples. It was observed that more realistic shapes reminiscent of signatures began to form following epoch 300.}
    \label{fig:gan-examples}
\end{figure}
\begin{figure}
    \centering
    \includegraphics[scale=0.75]{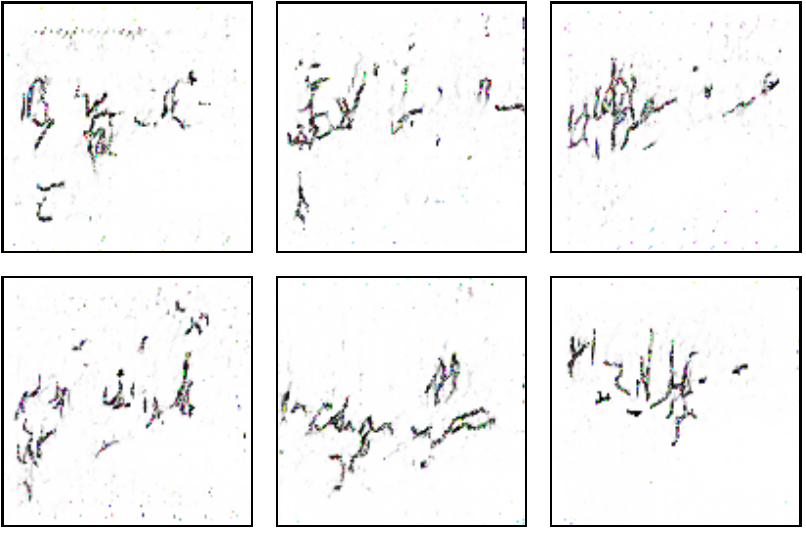}
    \caption{Examples of late generalisations by the GAN conditioned with the ``Real" class label.}
    \label{fig:late-examples}
\end{figure}
Figures \ref{fig:gen_chart} and \ref{fig:disc_chart} show the generator and discriminator losses for the Conditional GAN, respectively. Mode collapse did not occur with the chosen hyperparameters, and it was observed that relatively little change in the outputs occurred beyond epoch 1000. Realistic images reminiscent of signatures were also observed to form following the first 300 epochs, which can be seen in Figure \ref{fig:gan-examples}. Figure \ref{fig:late-examples} shows examples of how the GAN generalises ``Real" signatures within the latter stages of training. In some examples letters can be observed (e.g. the capital B in the first output), but it seems that the generator produces generalised clouds of pixels which can fool the discriminator.
\begin{figure}
    \centering
    \includegraphics[scale=0.85]{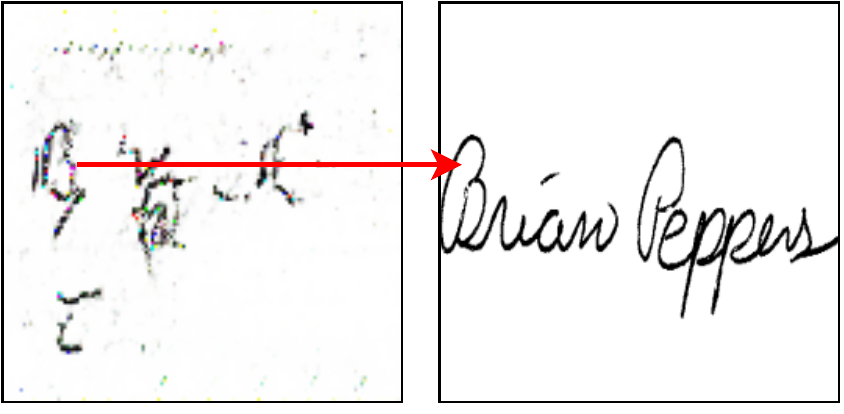}
    \caption{A possible GAN replication of a human-written letter compared to a similar sample from the dataset.}
    \label{fig:b-example}
\end{figure}
Figure \ref{fig:b-example} shows a comparison of the first GAN example and capital B within a signature in the data set.

\begin{table}[]
\centering
\caption{Successful attacks on writer-independent signature verification by the robots and GAN prior to tuning. }
\label{tab:attacks-results}
\begin{tabular}{@{}lrr@{}}
\toprule
\textbf{Method}             & \multicolumn{1}{l}{\textbf{Successful Attacks}} & \multicolumn{1}{l}{\textbf{Success (\%)}} \\ \midrule
\textit{\textbf{iDraw 2.0}} & 8/25                                           & 32                                                \\
\textit{\textbf{Line-us}}   & 6/25                                           & 24                                                \\
\textit{\textbf{cGAN}}      & 10/25                                          & 40                                                \\
\textit{\textbf{cGAN}}      & 297/1000                                       & 29.7                                              \\ \bottomrule
\end{tabular}
\end{table}
\begin{table}[]
\centering
\caption{Successful defence of writer-independent signature verification by robots and GAN after tuning. }
\label{tab:defense-results}
\begin{tabular}{@{}lrrr@{}}
\toprule
\textbf{Method}             & \multicolumn{1}{l}{\textbf{Successful Attacks}} & \multicolumn{1}{l}{\textbf{Success (\%)}} & \multicolumn{1}{l}{\textbf{Defense (\%)}} \\ \midrule
\textit{\textbf{iDraw 2.0}} & 2/25                                           & 8                                                 & +24                                       \\
\textit{\textbf{Line-us}}   & 3/25                                           & 12                                                & +12                                       \\
\textit{\textbf{cGAN}}      & 1/25                                           & 4                                                 & +36                                       \\
\textit{\textbf{cGAN}}      & 50/1000                                        & 5                                                 & +24.7                                     \\ \bottomrule
\end{tabular}
\end{table}

Table \ref{tab:attacks-results} shows the results of attacks on the system through robotic and GAN-based forgeries. All successful attacks are outside of the model baseline (87.12\%) and show the dangers of signature forgeries using such methods. Following tuning, a successful defence is then mounted against these approaches as shown by Table \ref{tab:results-dense}, wherein signature verification ability is improved by a minimum of 12\% and maximum of 24.7\%. Fine-tuning with examples brings the success of attack within the expected margin of error of the model. 

\section{Conclusion and Future Work}
\label{sec:conclusion}
This work has shown the dangers surrounding two adversarial attack methods on vision-based writer-independent signature verification models. Although the base model experienced over 87\% classification accuracy, successful attacks by robotic and GAN-based approaches were far above this margin of error. The Line-us and iDraw 2.0 robots were observed to fool the system 24\% and 32\% of the time, respectively. The Conditional GAN fooled the system 40\% of the time with a set of 25 signatures, and 29.7\% of the time with a set of 1000 signatures. In conclusion of this part of the study, the results have brought to the forefront information security issues when human versus human signatures forgeries are focused upon while a robot can achieve much more similar forgeries to the real human. Given this, biometric security could therefore be overcome by using these robots, which are low in cost and easily accessible. In the second part of the study, the verification model was fine-tuned with some examples generated by the approaches. Given exposure to robotic and GAN behaviours, the model could then prevent far more of these attacks; the accuracy of the attack was reduced by 12 and 24 percentage points for the Line-us (12\%) and iDraw 2.0 robots (8\%), respectively. The generative approach was successful only 4\% of the time for 25 images and 5\% of the time for a set of 1000 images. Given that these results were below the margin of error of the model, an acceptable defence has been mounted. 

In future, the study could be extended by allowing robots to generate a larger set of signatures. This is argued for by the two sets of results from the GAN (25 and 1000 images), which have a noticeable difference. In addition, further image classification approaches could be explored to form a stronger base model, such as attention modelling and residual information. 

To finally conclude, this study has shown that it is relatively easy to attack a signature verification model by generating forgeries physically with robots and digitally with generative approaches. This study has also shown a method to defend against such attacks through fine-tune transfer learning. The results after fine-tuning are effective, showing that such attacks can be prevented now, rather than after the first consumer robot-based forgery crime has been committed.

\bibliographystyle{IEEEtran}
\bibliography{references}

\end{document}